\documentclass[journal]{IEEEtran}

\ifCLASSINFOpdf
  \usepackage[pdftex]{graphicx}
  \graphicspath{{./figs/}}
  \DeclareGraphicsExtensions{.pdf,.jpeg,.png}
\else
\fi
\usepackage[hidelinks]{hyperref}
\usepackage[cmex10]{amsmath}
\usepackage{mathtools}
\usepackage{amssymb}
\usepackage{algorithm}
\usepackage{algpseudocode}

\usepackage{array}

\usepackage{booktabs}
\usepackage{multirow}
\usepackage{tabu}

\usepackage{biograph}        
\usepackage{afterpage}

\usepackage{cite}
\usepackage{multirow} 
\usepackage{rotating}

\usepackage{float}
\usepackage{color}
\usepackage{soul}

\usepackage{colortbl}

\def\por1{\partial}

\newcolumntype{S}{>{\centering\arraybackslash} m{.4\linewidth} }

\makeatletter
  \newcommand\tinyv{\@setfontsize\tinyv{5pt}{7}}
\makeatother

\newlength{\hspacephantom}
\begin{document}
\DeclareGraphicsExtensions{.pdf,.jpeg,.png}

\title{FSITM: A Feature Similarity Index\\ For Tone-Mapped Images}



\author{Hossein Ziaei Nafchi, Atena Shahkolaei, Reza Farrahi Moghaddam, \IEEEmembership{Member,~IEEE.},        \textnormal{and}
        Mohamed Cheriet, \IEEEmembership{Senior Member,~IEEE} 
\thanks{Copyright (c) 2014 IEEE. Personal use of this material is permitted. However, permission to use this material for any other purposes must be obtained from the IEEE by sending a request to pubs-permissions@ieee.org.}
\thanks{H. Ziaei Nafchi, A. Shahkolaei, R. Farrahi Moghaddam and M. Cheriet are with the Synchromedia Laboratory for Multimedia Communication in Telepresence,
\'Ecole de technologie sup\'erieure, Montreal (QC), Canada H3C 1K3; Tel.: +1-514-396-8972; Fax: +1-514-396-8595;
Emails: hossein.zi@synchromedia.ca, atena.shahkolaei.1@ens.etsmtl.ca, rfarrahi@synchromedia.ca, imriss@ieee.org, mohamed.cheriet@etsmtl.ca}
\thanks{Manuscript received ? ?, ?; revised ? ?, ?.}}

\markboth{}%
{Online non local patch means (NLPM) method}
%


\maketitle

\begin{abstract}
In this work, based on the local phase information of images, an objective index, called the feature similarity index for tone-mapped images (FSITM), is proposed. To evaluate a tone mapping operator (TMO), the proposed index compares the locally weighted mean phase angle map of an original high dynamic range (HDR) to that of its associated tone-mapped image calculated using the output of the TMO method. In experiments on two standard databases, it is shown that the proposed FSITM method outperforms the state-of-the-art index, the tone mapped quality index (TMQI). In addition, a higher performance is obtained by combining the FSITM and TMQI indices. The MATLAB source code of the proposed metric(s) is available at https://www.mathworks.com/matlabcentral/fileexchange/59814.
\end{abstract}


%
\IEEEpeerreviewmaketitle


\maketitle

\section{Introduction}
\label{sec:intro}


\IEEEPARstart{T}{here}
is increasing interest in high dynamic range (HDR) images, HDR imaging systems, and HDR displays. The visual quality of high dynamic range images is vastly higher than that of conventional low-dynamic-range (LDR) images, and the significance of the move from LDR to HDR has been compared to the momentous move from black-and-white to color television \cite{book2010}. In this transition period, and to guarantee compatibility in the future, there has been a need to develop methodologies to convert an HDR image into its `best' LDR equivalent. For this conversion, tone mapping operators (TMOs) have attracted considerable interest. Tone-mapping operators have been used to convert HDR images into their LDR associated images for visibility purposes on non-HDR displays.

 Unfortunately, TMO methods perform differently, depending on the HDR image to be converted, which means that the best TMO method must be found for each individual case. A survey of various TMOs for HDR images and videos is provided in \cite{Yeganeh2013} and \cite{survey2013}. Traditionally, TMO performance has been evaluated subjectively. In \cite{ACM2005}, a subjective assessment was carried out using an HDR monitor. Mantiuk et al. \cite{HDR-VDP} propose an HDR visible difference predictor (HDR-VDP) to estimate the visibility differences of two HDR images, and this tool has also been extended to a dynamic range independent image quality assessment \cite{independent2008}. However, the authors did not arrive at an objective score, but instead evaluated the performance of the assessment tool on HDR displays. Although subjective assessment provides true and useful references, it is an expensive and time-consuming process. In contrast, the objective quality assessment of tone mapping images enables an automatic selection and parameter tuning of TMOs \cite{Yeganeh2010, Yeganeh2014}. Consequently, objective assessment of tone-mapping images, which is proportional to the subjective assessment of the images, is currently of great interest.  

Recently, an objective index, called the tone mapping quality index (TMQI) was proposed in \cite{Yeganeh2013} to objectively assess the quality of the individual LDR images produced by a TMO. The TMQI is based on combining an SSIM-motivated structural fidelity measure with a statistical naturalness:
\begin{equation}
  \ \text{TMQI}(H,L)=a[S(H,L)]^\alpha + (1-a)[N(L)]^\beta.
  \label{TMQI}
\end{equation}  
where $S$ and $N$ denote the structural fidelity and statistical
naturalness, respectively. $H$ and $L$ denote the HDR and LDR images. The parameters $\alpha$ and $\beta$ determine the sensitivities of the two factors, and $a$ ($0\leq a \leq 1$) adjusts their relative importance. Both $S$ and $N$ are upper bounded by 1, and so the TMQI is also upper bounded by 1 \cite{Yeganeh2014}. Although the TMQI clearly provides better assessment for tone-mapped images than the well-known image quality assessment metrics, like SSIM \cite{SSIM}, MS-SSIM \cite{MSSSIM}, and FSIM \cite{FSIM}, its performance is not perfect. Liu et al. \cite{Liu2014} replaced the pooling strategy of the structural fidelity map in the TMQI with various visual saliency-based strategies for better quality assessment of tone mapped images. They examined a number of visual saliency models and conclude that integrating saliency detection by combining simple priors (SDSP) into the TMQI provides better assessment capability than other saliency detection models. 

In this paper, we first propose a feature similarity index for tone-mapped images (FSITM) which is based on the phase information of images. It has been observed that phase information of images prevails its magnitude \cite{oppenheim}. Also, physiological evidence indicates that the human visual system responds strongly to points in an image where the phase information is highly ordered \cite{morrone}. Based on this assumption, several quality assessment metrics have been proposed \cite{FSIM, contrast2013, signal2013}. In \cite{FSIM}, the maximum moment of phase congruency covariance, which is an edge strength map, is used. Hassen et al. \cite{contrast2013} used local phase coherence for image sharpness assessment. Saha et al. \cite{signal2013} proposed an image quality assessment using phase deviation sensitive energy features. Unfortunately, these metrics do not provide a reliable assessment for tone mapped images.  

The FSITM images proposed in this paper uses the phase-derived feature type of the images in a different way from that proposed in \cite{FSIM, contrast2013, signal2013}. Our FSITM uses a locally weighted mean phase angle (LWMPA) \cite{kovesipc}, which is a feature map based on the local-phase. This phase-derived map is noise independent, and therefore there is no parameter to set for noise estimation. The proposed FSITM assesses both the appearance of the real world scene and the most pleasing image for human vision.

Given the FSITM and the TMQI, we also proposed a combined metric, FSITM\_TMQI, which provides much better assessment of tone-mapped images. In the experiments, we compare the objective scores of our proposed similarity indices (FSITM, FSITM\_TMQI), along with TMQI \cite{Yeganeh2013}, on two major datasets \cite{dataset1,dataset2}. 





\section{The proposed similarity index}
\label{proposed}
The proposed FSITM similarity index for tone-mapped images is based on a phase-derived feature map. As we mentioned before, phase-derived features have already been used successfully for quality assessment \cite{FSIM, contrast2013, signal2013}. However, their results for evaluating tone-mapped images is not reliable similar to other popular quality assessment metrics like the SSIM and its variations \cite{SSIM, MSSSIM}. For this reason, we use the locally weighted mean phase angle (LWMPA) map in this paper, because it is a feature that is robust with respect to noise. Below, we briefly describe the theory and formulation of the LWMPA, and then discuss our proposed similarity index which is based on this feature map.

Let \(M_{\rho r}^e\) and \(M_{\rho r}^o\), which are known in the literature as quadratic pairs, denote the even symmetric and odd symmetric log-Gabor wavelets at a scale $\rho$ and orientation $r$ \cite{ivc2011}. By considering $f(\textbf{x})$ as a two-dimensional signal on the two-dimensional domain of $\textbf{x}$, the response of each quadratic pair of filters at each image point $\textbf{x}$ forms a response vector by convolving with $f(\textbf{x})$:
\begin{equation}
  \ \Big[e_{\rho r}(\textbf{x}),o_{\rho r}(\textbf{x})\Big]=\Big[f(\textbf{x})*M_{\rho r}^e,f(\textbf{x})*M_{\rho r}^o\Big]~.
  \label{response}
\end{equation}
where the values \(e_{\rho r}(\textbf{x})\) and \(o_{\rho r}(\textbf{x})\) are real and imaginary parts of a complex-valued wavelet response at a scale $\rho$ and an orientation $r$. We can now compute the local phase \(\phi_{\rho r}(\textbf{x})\) of the transform at a given wavelet scale \(\rho\) and orientation $r$:
\begin{equation}
  \ \phi_{\rho r}(\textbf{x})=\text{arctan2}\Big(o_{\rho r}(\textbf{x}),e_{\rho r}(\textbf{x})\Big) ~,
  \label{localphase}
\end{equation}
where arctan2(x,y) $=$ 2arctan$\frac{x}{\sqrt{x^2+y^2}+y}$.
The locally-weighted mean phase angle $\text{ph}(\textbf{x})$ is obtained using the summation of all filter responses over all the possible orientations and scales:
\begin{equation}
  \ \text{ph}(\textbf{x}) = \text{arctan2}\Big[\sum_{\rho ,r} e_{\rho r}(\textbf{x}),\sum_{\rho ,r} o_{\rho r}(\textbf{x})\Big] ~ .
 \label{LWMPA}
\end{equation}

The pixels of $\text{ph}(\textbf{x})$ take values between $-\pi/2$ (a dark line), $+\pi/2$ (a bright line), and 0 for steps. This classification of step and line features has been further studied in \cite{kovesistep}.

\begin{figure*}[htb]
\small
\begin{minipage}[b]{0.24\linewidth}
  \centering
  \centerline{\includegraphics[height=3.2cm]{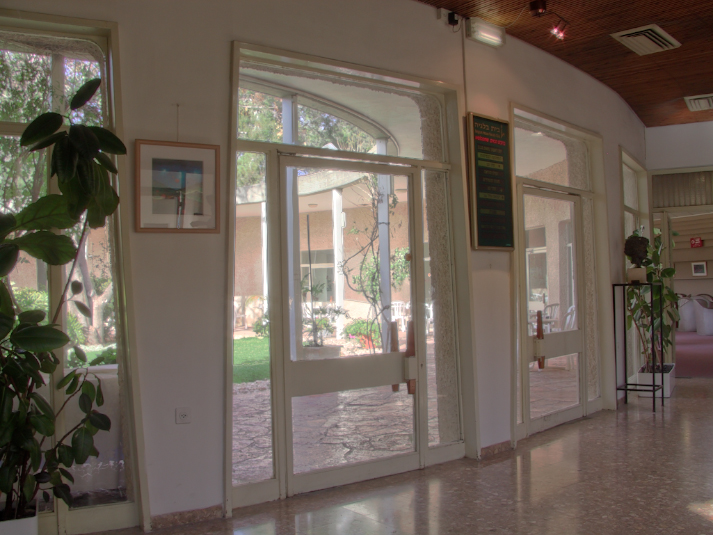}}
\centerline{(b) subjective score = 2.00}\medskip
\centerline{TMQI=0.9191, FSITM$^R$=0.8355}\medskip
\end{minipage}
\begin{minipage}[b]{.24\linewidth}
  \centering
  \centerline{\includegraphics[height=3.2cm]{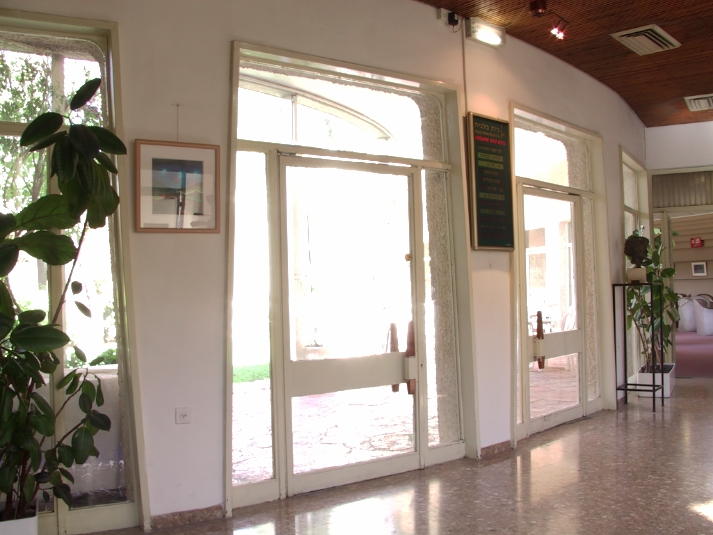}}
\centerline{(b) subjective score = 5.95}\medskip
\centerline{TMQI=0.8800, FSITM$^R$=0.7825}\medskip
\end{minipage}
\begin{minipage}[b]{0.24\linewidth}
  \centering
  \centerline{\includegraphics[height=3.2cm]{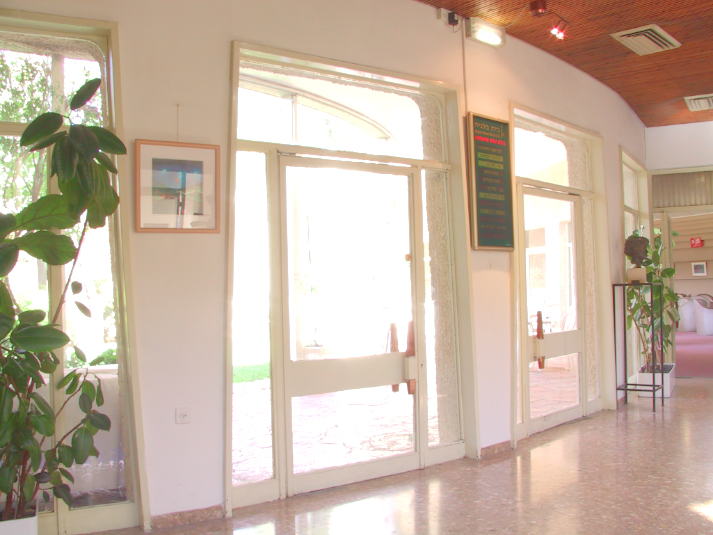}}
\centerline{(c) subjective score = 6.65}\medskip
\centerline{TMQI=0.7673, FSITM$^R$=0.7808}\medskip
\end{minipage}
\begin{minipage}[b]{0.24\linewidth}
  \centering
  \centerline{\includegraphics[height=3.2cm]{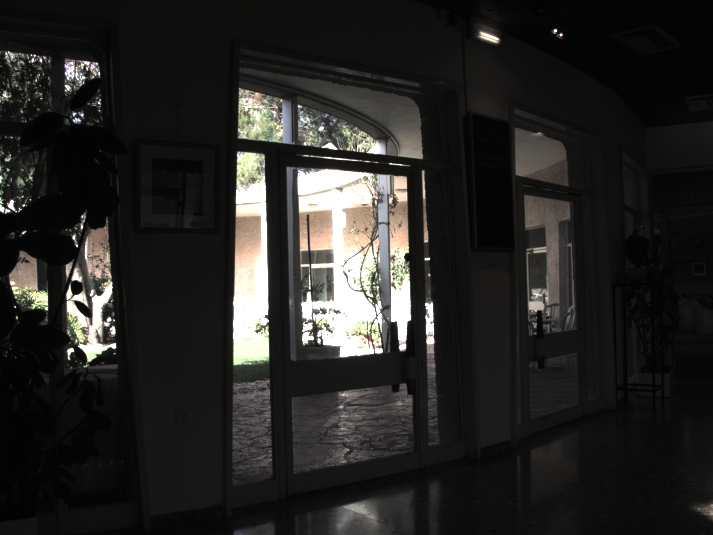}}
\centerline{(d) subjective score = 7.8}\medskip
\centerline{TMQI=0.7622, FSITM$^R$= 0.7514}\medskip
\end{minipage}
\\ \\
\begin{minipage}[b]{0.24\linewidth}
  \centering
  \centerline{\includegraphics[height=3.2cm]{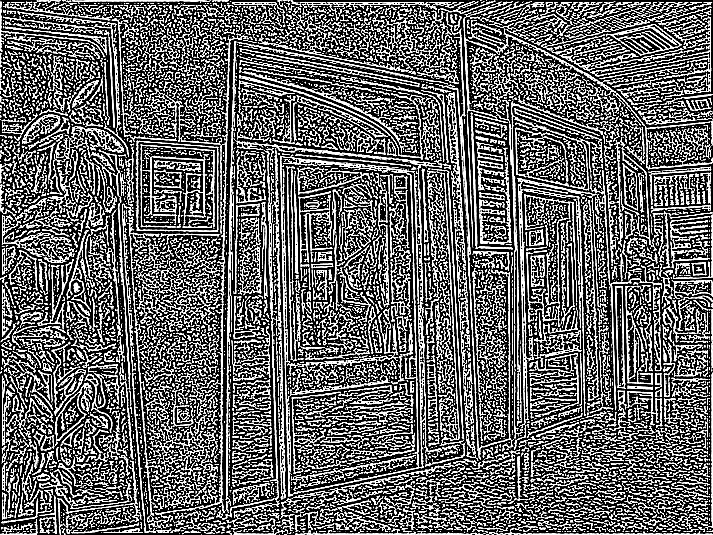}}
\centerline{(e)}\medskip
\end{minipage}
\begin{minipage}[b]{.24\linewidth}
  \centering
  \centerline{\includegraphics[height=3.2cm]{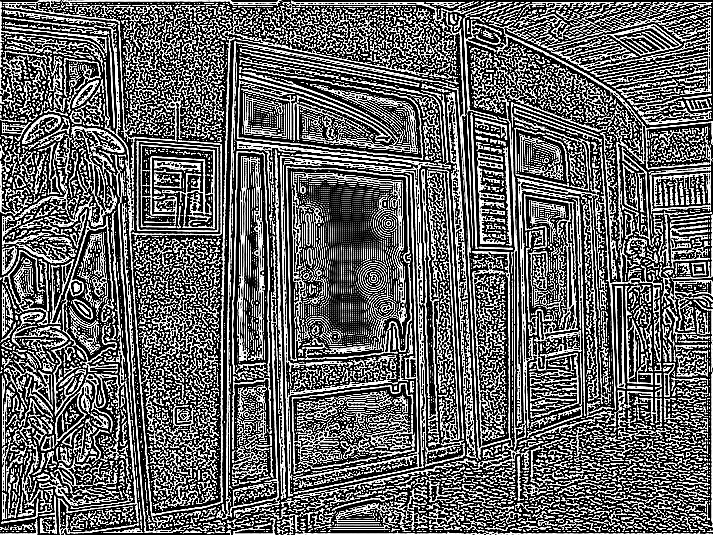}}
\centerline{(f)}\medskip
\end{minipage}
\begin{minipage}[b]{0.24\linewidth}
  \centering
  \centerline{\includegraphics[height=3.2cm]{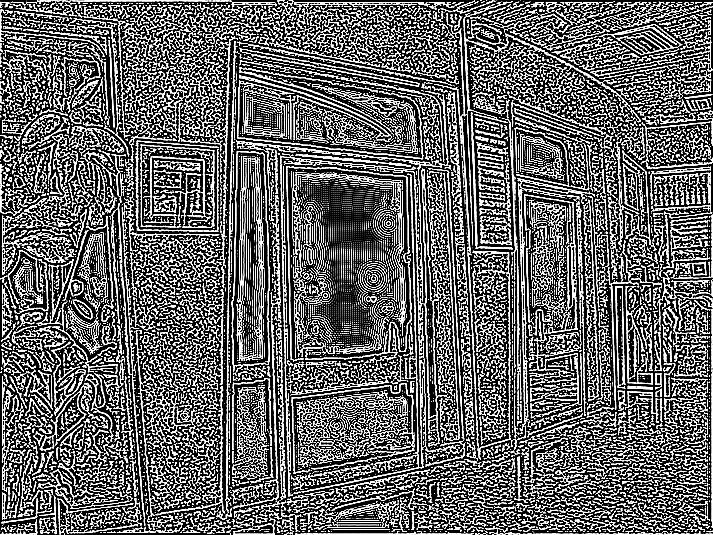}}
\centerline{(g)}\medskip
\end{minipage}
\begin{minipage}[b]{0.24\linewidth}
  \centering
  \centerline{\includegraphics[height=3.2cm]{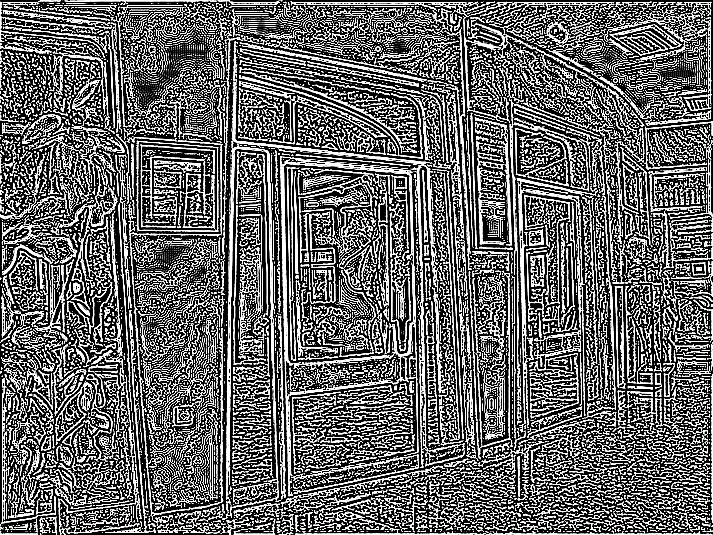}}
\centerline{(g)}\medskip
\end{minipage}
\caption{(a)-(d) Some LDR images using different TMOs \cite{Yeganeh2013}, along with their corresponding TMQI and FSITM scores for each. (e)-(g) The associated LWMPA maps of their red channel.}
\label{fig:out1}
\end{figure*}

There are a few parameters to be considered in the calculation of $\text{ph}(\textbf{x})$. In our set of experiments, we determine the best fixed values for this operation (see section \ref{experiments}). Unlike the phase-derived edge map and local phase that are used in other research \cite{FSIM, contrast2013}, the locally weighted mean phase angle $\text{ph}(\textbf{x})$ provides a good representation of image features, including the edges and shapes of objects. Since $\text{ph}(\textbf{x})$ indicates both dark and bright lines, it can be used to assess color changes, which is a popular feature of the TMOs. Moreover, the LWMPA is noise-independent, unlike the phase derived features used in \cite{FSIM, contrast2013, signal2013}, which are sensitive to noise, and therefore require an estimation of the noise. Some examples of $\text{ph}(\textbf{x})$ outputs are shown in Fig. \ref{fig:out1}. 

We use only $\text{ph}(\textbf{x})$ to calculate the FSITM. First, the HDR ($\text{H}$) image is converted into its LDR ($\text{L}$) by simply taking logarithm of its values ($\text{log(H)}$). This rough LDR image is used as one of the reference images for computing the FSITM. Another reference image is the HDR image itself. The details of the FSITM calculations are provided below.

Given the input images H and L, and $\text{LogH=log(H)}$ image, the $\text{ph}(\textbf{x})$ for each channel C of these three images is calculated using equation (\ref{LWMPA}). The FSITM is based on the simple fact that the features in the two corresponding channels should have remained the same in their $\text{ph}(\textbf{x})$ maps. The FSITM is equal to 1 if all the feature types are the same, and 0 if they are all different. First, we define the feature similarity index for a channel C used in calculation of the FSITM:
\begin{equation}
  \ \text{F}^C(I_1,I_2) = {|P_1^{C}(\textbf{x}) ~\cap~ P_2^{C}(\textbf{x})|} ~/~ ({row \times col}) ,
 \label{FSI}
\end{equation}
where P(\textbf{x}) denotes a binary image of ph(\textbf{x}):
\begin{equation}
  \ \text{P(\textbf{x})} = U\big(\text{ph(\textbf{x})}\big) ,
 \label{FSI}
\end{equation}
where $U(\cdotp)$ is the unit-step function. For the case of tone-mapped images, the FSITM for a channel C is defined as:
\begin{equation}
  \ \text{FSITM}^C = \alpha F^C(H, L) + (1-\alpha) F^C(\text{LogH}, L) .
 \label{FSI}
\end{equation}
where $\alpha$ ($0 \leq \alpha \leq 1$), controls the impact factor of H and LogH in the calculation of the FSITM. Algorithm \ref{alg1} lists all the steps in the process of calculating our proposed FSITM.

\begin{algorithm}[t!]
\small
\caption{The feature similarity index for tone-mapped images (FSITM).
} \label{alg1}
\begin{algorithmic}[1]
\Procedure{FSITM}{\textbf{H}, \textbf{L}, \textbf{C}}
\textbf{Start}
\State \textbf{H}: HDR,	~\textbf{L}: LDR, ~\textbf{C}$\in$\{R, G, B\}
\State \textbf{LogH} $=log$(\textbf{H}); 
\State Calculate $\text{ph}(\textbf{x})$ for \textbf{C} channel of images \textbf{L}, \textbf{H} and \textbf{LogH}. 
\State $\text{FSITM}^C = \alpha F^C(H, L) + (1-\alpha) F^C(logH, L)$;\\
\Return FSITM$^C$
\EndProcedure
\end{algorithmic}
\end{algorithm}

We also found that combining the FSITM and the TMQI provides a better assessment of tone-mapped images. Therefore, we proposed a combined index of the FSITM and the TMQI based on the following equation:
\begin{equation}
  \ \text{FSITM}^C\text{\_TMQI} = (\text{FSITM}^C + \text{TMQI})~/~{2}
 \label{comb1}
\end{equation}

In most of the cases, the different properties of these two indices cause them to moderate similarity estimation mistakes of each other. 

\section{Experimental results}
\label{experiments}

\begin{table*}[!t]

\centering
\caption{Performance comparison of the proposed quality indices and TMQI \cite{Yeganeh2013, dataset1} on the dataset A introduced in \cite{Yeganeh2013, dataset1}.}

\begin{tabular}{|c|c|c|c|c|c|c|c|}
\hline
\multicolumn{8}{|c|}{SRCC}                                                                                       \\ \hline
Index   & TMQI \cite{Yeganeh2013, dataset1} & FSITM$^\text{R}$ & FSITM$^\text{R}$\_TMQI & FSITM$^\text{G}$ & FSITM$^\text{G}$\_TMQI & FSITM$^\text{B}$        & FSITM$^\text{B}$\_TMQI \\ \hline
Min     & 0.6826   & 0.6190   & \textbf{0.7143} & 0.5476   & \textbf{0.7143} & 0.1796          & 0.5509          \\ \hline
Median  & 0.7857   & 0.8095   & \textbf{0.8571} & 0.8333   & \textbf{0.8571} & \textbf{0.8571} & \textbf{0.8571} \\ \hline
Average & 0.8058   & 0.8145   & \textbf{0.8559} & 0.8178   & 0.8424          & 0.7183          & 0.8097          \\ \hline
STD     & 0.1051   & 0.1214   & \textbf{0.0863} & 0.1310   & \textbf{0.0886} & 0.2536          & 0.1229          \\ \hline
\multicolumn{8}{|c|}{KRCC}                                                                                       \\ \hline
Min     & 0.5455   & 0.5000   & \textbf{0.5714} & 0.3571   & \textbf{0.5714} & 0.2143          & 0.4001          \\ \hline
Median  & 0.6429   & 0.7143   & 0.7143          & 0.7143   & \textbf{0.7857} & 0.7143          & 0.7143          \\ \hline
Average & 0.6840   & 0.7126   & \textbf{0.7508} & 0.6935   & 0.7317          & 0.5979          & 0.6838          \\ \hline
STD     & 0.1221   & 0.1423   & \textbf{0.1083} & 0.1711   & \textbf{0.1078} & 0.2711          & 0.1436          \\ \hline
\end{tabular}
\label{table1}
\end{table*}

\begin{table*}[!t]

\centering
\caption{Performance comparison of the proposed quality indices and TMQI \cite{Yeganeh2013, dataset1} on the dataset B introduced in \cite{dataset2, Cadic2008}.}

\begin{tabular}{|c|c|c|c|c|c|c|c|}
\hline
\multicolumn{8}{|c|}{SRCC}                                                                                \\ \hline
Index   & TMQI \cite{Yeganeh2013, dataset1} & FSITM$^\text{R}$ & FSITM$^\text{R}$\_TMQI & FSITM$^\text{G}$ & FSITM$^\text{G}$\_TMQI & FSITM$^\text{B}$ & FSITM$^\text{B}$\_TMQI \\ \hline
Min     & 0.7198   & 0.7363   & 0.8901          & 0.7692   & \textbf{0.9231} & 0.7637   & 0.8462          \\ \hline
Average & 0.7985   & 0.7692   & 0.9102          & 0.8461   & \textbf{0.9267} & 0.8241   & 0.8901          \\ \hline
\multicolumn{8}{|c|}{KRCC}                                                                                \\ \hline
Min     & 0.5385   & 0.5897   & 0.6923          & 0.6154   & \textbf{0.7692} & 0.5385   & 0.6410          \\ \hline
Average & 0.6410   & 0.6410   & 0.7692          & 0.7265   & \textbf{0.8119} & 0.6410   & 0.7264          \\ \hline
\end{tabular}
\label{table2}
\end{table*}

To evaluate the proposed FSITM index, we used the dataset (dataset A) introduced in \cite{Yeganeh2013} and \cite{dataset2}. The first dataset contains 15 HDR images, along with 8 LDR images for each HDR image. The HDR images were produced using different TMOs. The quality of LDRs is ranked from 1 (best quality) to 8 (worst quality). The ranks were obtained based on a subjective assessment of 20 individuals. The second HDR dataset (dataset B) used is also available along with subjective ranks for LDR images \cite{dataset2, Cadic2008}. That dataset contains three HDR images, and 14 LDR images for each HDR image. 

To objectively evaluate the performance of the various similarity indices considered in our experiments, we use the Spearman rank-order correlation coefficient (SRCC) and the Kendall rank-order correlation coefficient (KRCC) metrics.

The proposed similarity indices (FSITM$^C$, FSITM$^C$\_TMQI) are compared with the TMQI \cite{Yeganeh2013}. The results are listed in Tables \ref{table1} and \ref{table2}. The performance of the TMQI is listed based on the scores obtained by running the Matlab source code provided by Yeganeh and Wang in \cite{dataset1}. The FSITM$^G$ outperforms the TMQI in terms of SRCC and KRCC for both datasets. In general, there is less variation in TMQI performance than in FSITM performance. In contrast, the FSITM$^R$\_TMQI and FSITM$^G$\_TMQI are more robust, and also they outperform the FSITM and TMQI in terms of both the SRCC and KRCC scores.  

It is worth to report the available results of other indices which have recently been proposed in the literature \cite{Liu2014, blind2014}. In \cite{Liu2014}, the authors reported the SRCC performance of their proposed index for the dataset A \cite{dataset1}. Their minimum and average SRCC performance is 0.6905 and 0.8408, respectively. Their standard deviation of SRCC scores is reported as 0.0907. For the same dataset, the median performance of the ref. \cite{blind2014} is reported as follows: SRCC$=$0.8106 and KRCC$=$0.5865.

A number of parameters impact the quality of the locally weighted mean phase angle $\text{ph}(\textbf{x})$, namely the number of filter scales $N_\rho$, the wavelength of the smallest scale filter wLen, and the scaling factor between successive filters  mult. In the experiments, these parameters were set to $N_\rho = 2$, $wLen = 2$, and $mult = 2$ for the LogH image, while they were set to $N_\rho = 2$, $wLen = 8$, and $mult = 8$ for the original HDR image. The rational for using two different set of parameters is that the size of the image features could be different. Overall, it is the three parameters of $\text{ph}(\textbf{x})$ along with the value of $\alpha$ that influence the performance of the proposed indices.  

In this work, we only used the original HDR image and its logarithm image LogH. It is worth mentioning that we have tried the same strategy used in defining FSITM in RGB color space in other color spaces, such as Lab and Yxy color spaces. However, we did not get a good performance.    

We evaluated the run time of the FSITM and the TMQI as follows: our experiments were performed on a Core i7 3.4 GHz CPU with 16 GB of RAM. The FSITM algorithm was implemented in MATLAB 2012b running on Windows 7. The TMQI and the FSITM took 1.95 and 3.36 seconds respectively to assess images of size 1200$\times$1600, while the run time for the FSITM$^C$\_TMQI is simply obtained by adding the TMQI and FSITM$^C$ run-times.


%



\section{Conclusion}
\label{conclusion}
We have proposed an objective index, called the feature similarity index for tone-mapped images (FSITM), which is based on the local phase similarity of the original HDR and the target converted LDR image. Unlike other studies in which different phase-derived feature maps are used, we have used the locally weighted mean phase angle, which is a robust and noise-independent feature map. The performance of the proposed similarity index is compared with the state-of-the-art TMQI on two datasets, and has been found to be promising. The proposed FSITM and the TMQI have been then combined to obtain a more accurate quality assessment. Further studies are required to develop more comprehensive HDR datasets, along with their subjective scores. Such datasets would allow us to develop better performing indices.

\section*{Acknowledgments}
The authors thank the NSERC of Canada for their financial support under Grants RGPDD 451272-13 and RGPIN 138344-14.


\bibliographystyle{IEEEtran}
\bibliography{egbib2}   

%






\end{document}